\documentclass[conference]{IEEEtran}
\IEEEoverridecommandlockouts

\usepackage{comment}
\usepackage{caption}
\usepackage{stfloats} 
\usepackage{booktabs}
\usepackage{diagbox}
\usepackage{tabularx}
\usepackage{multirow} 
\usepackage{pifont}

\usepackage{varwidth}
\usepackage{amsmath,amssymb,amsfonts,graphicx}
\usepackage{makecell}
\usepackage{hyperref}
\usepackage{cite}
\usepackage{graphicx,epstopdf,caption,url}  
\usepackage{amssymb} 
\usepackage{array}   
\usepackage{arydshln} 
\usepackage{graphics}
\usepackage{color}   
\usepackage{graphicx}

\begin{document}
\title{Large Language Models in Education: \\ Vision and Opportunities}

\author{Wensheng Gan$ ^{1,2*}$\thanks{\IEEEauthorrefmark{1}Corresponding author: wsgan001@gmail.com. Please cite: W. Gan, Z. Qi, J. Wu, and J. C. W. Lin, “Large Language Models in Education: Vision and Opportunities,” in \textit{IEEE International Conference on Big Data}, pp. 1–10, 2023.}, Zhenlian Qi$ ^{3}$, Jiayang Wu$ ^{1}$, Jerry Chun-Wei Lin$ ^{4}$ \\ 	\\
	$ ^{1} $Jinan University, Guangzhou 510632, China\\
	$ ^{2} $Pazhou Lab, Guangzhou 510330, China\\
        $ ^{3} $Guangdong Eco-Engineering Polytechnic, Guangzhou 510520, China\\
        $ ^{4} $ Silesian University of Technology, 44-100 Gliwice, Poland.\\
}

\maketitle

\begin{abstract}
    With the rapid development of artificial intelligence technology, large language models (LLMs) have become a hot research topic. Education plays an important role in human social development and progress. Traditional education faces challenges such as individual student differences, insufficient allocation of teaching resources, and assessment of teaching effectiveness. Therefore, the applications of LLMs in the field of digital/smart education have broad prospects. The research on educational large models (EduLLMs) is constantly evolving, providing new methods and approaches to achieve personalized learning, intelligent tutoring, and educational assessment goals, thereby improving the quality of education and the learning experience.  This article aims to investigate and summarize the application of LLMs in smart education. It first introduces the research background and motivation of LLMs and explains the essence of LLMs. It then discusses the relationship between digital education and EduLLMs and summarizes the current research status of educational large models. The main contributions are the systematic summary and vision of the research background, motivation, and application of large models for education (LLM4Edu). By reviewing existing research, this article provides guidance and insights for educators, researchers, and policy-makers to gain a deep understanding of the potential and challenges of LLM4Edu. It further provides guidance for further advancing the development and application of LLM4Edu, while still facing technical, ethical, and practical challenges requiring further research and exploration.
\end{abstract}


\IEEEpeerreviewmaketitle

\section{Introduction}  \label{sec:introduction}

With the rapid development of big data \cite{sun2022big,sun2023internet}, artificial intelligence, and Web 3.0 \cite{wan2023web3,gan2023web}, large language models (LLMs) \cite{zhao2023survey,gan2023large,gan2023model,zeng2023large} have become a research hotspot. LLMs are deep learning models that learn the underlying patterns and rules of language by training on large-scale corpora. They possess powerful capabilities in generating and understanding natural language and have been widely applied in natural language processing (NLP) \cite{wu2023ai}, machine translation \cite{xiao2023survey}, dialogue systems \cite{ziems2021moral}, AI-generated content (AIGC) \cite{wu2023ai}, social cognitive computing, among other fields. Education is a significant domain that plays a crucial role in the development and progress of human society. Traditional educational models face challenges such as individual differences among students, insufficient allocation of teaching resources, and the assessment of teaching effectiveness \cite{aldhafeeri2022effectiveness}. Therefore, incorporating LLMs into the field of education holds the potential to provide support for personalized learning \cite{lin2022metaverse}, intelligent tutoring, adaptive assessment \cite{mcnamara2023istart}, and other aspects, thereby improving the quality of education and the learning experience.

In the digital era, the education field currently faces various challenges \cite{lin2022metaverse}, including low student engagement \cite{kristianto2023offline} and unequal distribution of teaching resources \cite{smith2016widening}. Traditional classroom teaching struggles to meet the personalized needs of different students. LLMs, as powerful natural language processing tools, have the potential to revolutionize traditional teaching models by enabling personalized learning and intelligent tutoring. Furthermore, with the advent of the big data era, the education field has accumulated a vast amount of learning data \cite{piety2014educational}. Utilizing this data for in-depth analysis and mining can reveal learners' patterns \cite{vermunt2017learning}, evaluate learning outcomes \cite{aziz2012evaluation}, and provide personalized recommendations \cite{fang2021personalized,bhargava2022commonsense}. LLMs have advantages in processing and analyzing large-scale data, making their application in the education field capable of providing deeper learning support and personalized education.

Large models refer to models with a massive number of parameters and computational capabilities \cite{kasneci2023chatgpt}. LLMs are one type of large models, often involving billions of parameters. The essence of large models lies in their ability to handle complex tasks and large-scale data, enabling them to learn richer language patterns and knowledge representations \cite{bhargava2022commonsense}. This makes large models highly applicable in the field of education. Smart/intelligent education refers to the provision of personalized, adaptive, and intelligent educational services through the utilization of technologies such as artificial intelligence and big data. For smart education, educational large models (EduLLMs) refer to educational application models based on LLMs. By learning from extensive educational data and corpora, EduLLMs can provide personalized learning support \cite{raj2022systematic}, intelligent tutoring \cite{wang2023unified}, and educational assessment capabilities to students \cite{rudolph2023chatgpt}. The research status of EduLLMs demonstrates significant potential and opportunities. Firstly, EduLLMs can identify students' learning patterns and characteristics by learning from massive educational data, enabling the provision of personalized learning support and recommendations for educational resources. Secondly, EduLLMs can be applied to intelligent tutoring, providing real-time problem-solving, learning advice, and academic guidance through dialogue and interaction with students. Moreover, EduLLMs have the potential for educational assessment, automatically evaluating students' knowledge mastery, learning outcomes, and expressive abilities, thereby providing more comprehensive student evaluation and teaching feedback to educators.

However, the research on LLM4Edu still faces challenges and issues. Firstly, social cognitive learning is challenging in LLM4Edu. Data privacy and security are crucial considerations to ensure the protection of students' personal information \cite{marshall2022implementing}. The interpretability and fairness of LLM4Edu are also focal points \cite{kizilcec2022algorithmic}, requiring the large models' decision-making processes to be interpretable and avoiding unfair biases caused by data. Moreover, the development and deployment of educational large models need to fully consider educational practices and teachers' professional knowledge to ensure the models are closely integrated with actual teaching \cite{lee2023rise}.

This paper is a systematic summary and analysis of the research background, motivation, and applications of educational large models. By reviewing existing research, we provide an in-depth understanding of the potential and challenges of educational large models for education practitioners, researchers, and policymakers, offering guidance and insights for further advancing the development and application of EduLLMs. The main contributions of this article are as follows: 

\begin{itemize}
    \item This paper first reviews the background of education, LLMs, and smart education, respectively. It then introduces the connection between LLMs and education and also discusses mart education (Section \ref{sec:background}).
    \item This paper provides an in-depth understanding of the key technologies of EduLLMs, including natural language processing (NLP), machine learning, data mining, computer vision, etc. (Section \ref{sec:technologies}).
    
    \item We also discuss how LLMs empower education from the perspective of various applications of education under LLMs (Section \ref{sec:applications}). It further exhibits several distinct characteristics of education under LLMs (Section \ref{sec:characteristics}).
    
    \item We also summarize the key points in EduLLMs, including training data and preprocessing, the training process, and integration with various technologies (Section \ref{sec:keys}). 
    
    \item Finally, we highlight some key challenges existing in LLM4Edu (Section \ref{sec:challenges}), and discuss potential future directions for LLM4Edu in more detail (Section \ref{sec:directions}).
\end{itemize}

\section{Education and LLMs} \label{sec:background}
\subsection{Background of Education}

Education is a conscious process of facilitating and guiding individual development \cite{zachary2022mentor}. It involves imparting knowledge, fostering skills, and shaping attitudes and values, with the aim of promoting holistic growth and self-realization in learners. The goal of education is to cultivate intellectual, emotional, moral, creative, and social adaptability in individuals, enabling them to make positive contributions to society.

Education takes various forms, including but not limited to:
\begin{itemize}
    \item \textbf{School education}: Traditional school education is the most common and widely accepted form, where students receive organized instruction from teachers and acquire knowledge and skills.

    \item \textbf{Online education}: With the advancement of digital technologies, the internet and online platforms provide new forms of education \cite{shunkov2022prospective}. Students can engage in learning through online courses, distance education, and other digital avenues.

    \item \textbf{Community education}: It refers to educational activities conducted within a community, providing specific training and learning opportunities to meet the educational needs of community members.

    \item \textbf{Self-directed learning}: Learning is the key to education. Self-directed learning emphasizes the ability of students to explore and learn autonomously, acquiring knowledge and skills through self-motivation and self-management.
\end{itemize}

In general, education involves various roles, including but not limited to:
\begin{itemize}
    \item \textbf{Teachers}: Teachers have a core role in education. They are responsible for organizing, imparting knowledge, and guiding student learning and development.

    \item \textbf{Students}: Students are the recipients of education. They acquire knowledge and skills through learning and practice, aiming for personal development and growth.

    \item \textbf{Parents}: As an important supportive and guardianship role in education, they are concerned with their children's learning and development, providing necessary resources and environments.

    \item \textbf{Educational institutions}: Schools, universities, training organizations, and other educational institutions provide educational resources and environments, organizing and managing educational activities.

    \item \textbf{Government and society}: They play roles in education policy-making, resource allocation, and social support, providing necessary support and safeguards for education.
\end{itemize}

\subsection{Background of LLMs}

What is a large language model (LLM) \cite{zhao2023survey,kasneci2023chatgpt}?  What are its characteristics? What is the relationship between large models and AI, data science, and other interdisciplinary fields? What are the key technologies employed in large models? A LLM possesses powerful language generation and understanding capabilities. Its objective is to train on massive amounts of language data to learn the statistical patterns and semantic relationships within the language, to generate coherent and accurate text, and to understand and respond to human queries \cite{tang2023science}. Here are several characteristics of LLMs:

\textbf{1. Natural language generation}: LLMs can generate high-quality, coherent natural language text. They can understand the context and generate appropriate responses, articles, stories, and more based on input prompts or questions \cite{baidoo2023education}.

\textbf{2. Semantic understanding}: LLMs can comprehend the semantic relationships within human language, including vocabulary, syntax, and context \cite{weissweiler2022better}. They can parse and understand complex sentence structures, extract key information, and generate relevant responses.

\textbf{3. Context awareness}: LLMs can perform language understanding and generation based on context \cite{meng2022generating}. They can understand the history of a conversation and generate responses that are coherent and related to the context.

\textbf{4. Wide range of applications}: LLMs have extensive applications in natural language processing, virtual assistants \cite{agarwal2022chatbots}, intelligent customer service \cite{gao2022impact}, and intelligent writing \cite{salvagno2023can}, among others. They can provide language generation and understanding support for various tasks and scenarios.

\textbf{5. Continuous learning}: LLMs can continuously learn and update themselves by training on new data \cite{ertuugrul2023lifelong}. They can accumulate new language knowledge and patterns by learning from fresh data, improving their performance and capabilities.

Large models employ several key technologies. Here, we describe five of them in detail:

\textbf{1. Transformer model}: It serves as the foundational architecture for large models \cite{vaswani2017attention}. It utilizes self-attention mechanisms to handle the dependency relationships within input sequences \cite{shaw2018self}. It effectively captures long-range dependencies, enabling the model to better understand and generate text.

\textbf{2. Pre-training and fine-tuning}: Large models typically employ a two-stage approach of pre-training \cite{zoph2020rethinking} and fine-tuning \cite{howard2018universal}. In the pre-training stage, the model undergoes self-supervised learning using a large-scale unlabeled corpus to learn the statistical patterns and semantic relationships of language. In the fine-tuning stage, the model is further trained and adjusted using labeled task-specific data to adapt to specific task requirements.

\textbf{3. Large-scale datasets}: Large models require massive language datasets for training \cite{kandpal2023large}. These datasets often include text data from the internet, books, news articles, and more. The use of large-scale data provides abundant language inputs and enhances the model's generalization ability.

\textbf{4. High computational resources} \cite{zeng2023distributed}: Large models necessitate significant computational resources for training and inference. High-performance graphics processing units (GPUs) or specialized deep learning accelerators, such as TPUs, are commonly used to accelerate computations and achieve efficient model training and inference.

\textbf{5. Iterative optimization algorithms}: Large models are typically trained using iterative optimization algorithms such as stochastic gradient descent (SGD) \cite{jin2023convergence} and adaptive optimization algorithms like ADMA \cite{reyad2023modified}. These algorithms update the model's parameters through backpropagation, minimizing the loss function and optimizing the model's performance.

In addition to the aforementioned key technologies, research on large models also involves aspects such as scaling up model size \cite{kang2023scaling}, data handling and selection \cite{zhu2023unsupervised}, model compression and acceleration \cite{xu2023survey}, and more. With advancing technology, the application of large models in natural language processing, intelligent dialogues, text generation, and other fields will become more extensive and mature.

\subsection{Smart Education}

Smart education refers to an educational model that utilizes advanced information technology and the theories and methods of educational science to provide personalized, efficient, and innovative learning and teaching experiences. Its core idea is to leverage the advantages of information technology to offer intelligent and personalized learning environments and resources, thereby promoting students' comprehensive development and enhancing learning outcomes.

Smart education is closely related to artificial intelligence (AI) and LLMs \cite{bajaj2018smart}. AI is the scientific and engineering field that aims to simulate and mimic human intelligence, while LLMs are a type of deep learning model with the capability to handle large-scale data and complex tasks. Through the applications of AI and LLMs, smart education can achieve more accurate learning analysis and assessment, personalized learning support and guidance, automated learning resource recommendations, and innovative teaching methods. However, smart education currently faces several issues and challenges:

\begin{itemize}
    \item \textbf{Shift in roles for teachers and students}: Smart education involves transforming the roles of teachers and students from traditional transmitters and receivers of knowledge to collaborators and explorers \cite{hampel2001steam}. This requires teachers to possess new teaching philosophies and skills to adapt to and guide students in the learning approaches and needs within a smart education environment.

    \item \textbf{Data privacy and security} \cite{chen2022metaverse,chen2023privacy}: Smart education involves the collection and analysis of large amounts of student data to provide personalized learning support and assessment. However, this raises concerns about student privacy and data security \cite{may2011using}. It is crucial to establish robust data management and protection mechanisms to ensure the safety and lawful use of student data.

    \item \textbf{Technological infrastructure and resources}: Implementing smart education requires adequate technological infrastructure and resource support, including network connectivity, computing devices, educational software, etc. However, some regions and schools may face challenges regarding technological conditions and resource scarcity, limiting the widespread adoption and application of smart education.

    \item \textbf{Ethical and moral issues}: The application of smart education raises ethical and moral questions, such as data privacy, algorithm bias, and fairness in artificial intelligence. It is necessary to establish guidelines and regulations to ensure that the application of smart education not only yields educational benefits but also adheres to ethical principles and social fairness.

    \item \textbf{Balancing personalization and social equity}: Smart education aims for personalized learning support, but excessive reliance on personalization may widen the gaps between learners. It is essential to strike a balance between personalization and social equity, ensuring that the application of smart education does not exacerbate educational inequalities but instead provides equal learning opportunities for all learners.
\end{itemize}

In conclusion, smart education refers to an educational model that utilizes advanced information technology and the theories and methods of educational science to provide personalized, efficient, and innovative learning and teaching experiences. It is closely related to AI and large models. However, unlike mere technological applications, smart education also involves a range of issues and challenges, including the transformation of teacher roles, data privacy and security, technological infrastructure and resources, ethical and moral concerns, balancing personalization and social equity, and innovation in educational content and assessment systems. Addressing these issues and promoting the sustainable development of education requires collaborative efforts from the education sector, technology industry, and society as a whole.

\subsection{LLMs for Education}

Large models have close relationships with artificial intelligence, data science, and other interdisciplinary fields. Large models are an important research direction within the field of artificial intelligence. They use deep learning and large-scale data training methods to simulate human language capabilities and achieve natural language processing tasks. In the field of data science, large models can be applied to tasks such as text mining, sentiment analysis, machine translation, and extracting valuable information from text data. Furthermore, large models involve computer science, machine learning, cognitive science, and other interdisciplinary fields. Through the study of language and intelligence, they drive the cross-fertilization and development between these disciplines. 

In recent years, the emergence of LLMs, such as GPT-3, has sparked widespread attention and discussion. LLMs are AI technologies based on deep learning that possess powerful language generation and understanding capabilities. At the same time, the field of education faces many challenges and opportunities, such as personalized learning, educational resource inequality, and instructional effectiveness assessment. As a result, the education sector has begun to explore how to integrate LLMs with education to enhance teaching quality and effectiveness. Here are the significance and several ongoing practical areas, which can be depicted in Fig. \ref{fig:edu}:

\begin{figure}[ht]
    \centering
    \includegraphics[clip,scale=0.37]{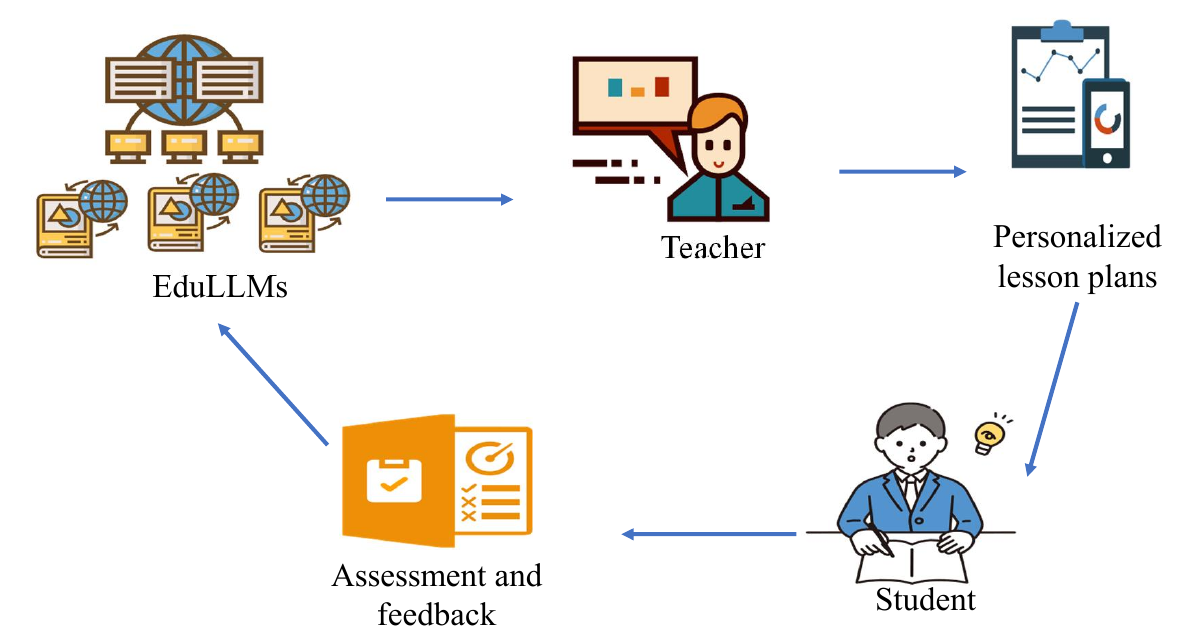}
    \caption{Architecture of LLMs for education (LLM4Edu).}
    \label{fig:edu}
\end{figure}

\textbf{1. Personalized learning}: Large models can provide personalized learning content and recommendations based on students' learning needs and interests. By analyzing students' learning data and behavioral patterns \cite{fournier2022pattern,gan2019survey}, large models can design unique learning paths and resources for each student \cite{herodotou2019large}, helping them learn and grow more efficiently.

\textbf{2. Instructional support tools}: LLMs can serve as assistants to teachers, providing intelligent instructional support tools and platforms \cite{filgueiras2023artificial}. Teachers can utilize the generated content and recommendations from LLMs to design teaching activities, monitor students' learning progress, and provide personalized teaching support.

\textbf{3. Educational assessment and feedback}: LLMs can analyze students' assignments, exams, and other learning data to provide assessment and feedback on their learning progress. By automatically generating comments and suggestions, LLMs can help teachers gain a more accurate understanding of students' learning achievements and challenges, and provide corresponding guidance and support.

\textbf{4. Educational resource and content creation}: LLMs can be used for the creation and generation of educational resources and content. They can generate teaching materials, exercises, case studies, and more based on instructional goals and needs, providing teachers with a rich array of resources and inspiration.

\section{Key Technologies for EduLLMs} \label{sec:technologies}

Educational LLMs involve several key technologies. Here are 10 key technologies related to educational large language models (EduLLMs), along with detailed descriptions for each:

\textbf{1. Natural language processing (NLP)}: NLP is one of the core technologies behind EduLLMs. It encompasses techniques such as text analysis, semantic understanding, and sentiment analysis, enabling the models to comprehend and process human language \cite{al2023survey}. NLP enables EduLLMs to understand student queries, generate language responses, and extract important information from text.

\textbf{2. Deep learning (DL}): DL is a branch of machine learning \cite{jordan2015machine} that involves constructing and training deep neural network models for learning and inference \cite{lecun2015deep}. EduLLMs often rely on deep learning architectures such as convolutional neural networks (CNNs) or recurrent neural networks (RNNs) to process and analyze educational data and generate meaningful outputs. Many DL techniques have been developed.

\textbf{3. Reinforcement learning (RL)} \cite{kaelbling1996reinforcement}: RL trains an agent to make decisions through trial and error and reward mechanisms. In EduLLMs, reinforcement learning can be employed to optimize model responses and recommendations, allowing the models to adjust based on student feedback and outcomes to provide more accurate and effective learning support \cite{carta2023grounding}.

\textbf{4. Data mining (DM) \cite{gan2020huopm,gan2021survey}}: DM is the process of extracting useful information and patterns from large datasets. EduLLMs can utilize data mining techniques to discover student learning patterns, behavior trends, and knowledge gaps, providing the foundation for personalized learning and offering insights for educational research.

\textbf{5. Computer vision (CV)}: The powerful CV technologies enable computers to understand and interpret images and videos. In education, EduLLMs can employ computer vision techniques to analyze students' facial expressions, postures, and behaviors, providing more accurate emotion analysis and learning feedback \cite{thomas2017predicting}.

\textbf{6. Speech recognition and synthesis}: Speech recognition technology converts speech into text, while speech synthesis technology converts text into speech. EduLLMs can utilize these technologies to engage in speech interactions with students, offering support for oral practice, speech assessment, and pronunciation correction \cite{wong2022leveraging}.

\textbf{7. Multimodal learning} \cite{wu2023multimodal}: It involves the fusion of various sensors and data sources, such as text, images, audio, and video. EduLLMs can process and analyze multimodal data to gain a more comprehensive understanding of students' learning situations and needs \cite{martinez2020data}.

\textbf{8. Personalized recommendation systems}: They utilize ML and DM techniques to provide students with personalized learning resources and suggestions based on their interests, learning history, and learning styles \cite{li2023prompt}. EduLLMs can play a significant role in personalized recommendation systems, leveraging student data and behavior patterns to recommend suitable learning materials, courses, and activities.

Therefore, the combination of these key technologies enables EduLLMs to offer personalized, adaptive, and targeted educational support. The applications foster innovation in education, improving learning outcomes and teaching quality. However, these applications of EduLLMs also face challenges, such as privacy protection, data bias, and algorithm transparency. These need to be appropriately addressed in technological development and practical implementation.

\section{LLM-empowered Education} 
\subsection{Applications of Education under LLMs}
\label{sec:applications}

Possible applications of LLMs for education can be found in various educational scenarios, providing personalized learning, teaching assistance, and educational research support. Here are 12 potential application scenarios of LLM4Edu, along with specific descriptions and examples, as shown in Table \ref{table:examples}:

\begin{table*}[ht]
    \caption{Several applications of LLM4Edu}
    \label{table:examples}
    \centering
    \begin{tabularx}{0.82\textwidth}{|>{\centering\arraybackslash}X|p{8cm}|}
        \hline
        \multicolumn{1}{|c|}{\textbf{Function}} & \multicolumn{1}{c|}{\textbf{Description}}  \\
        \hline
        \textbf{Learning assistance tools} & Provide support in problem-solving, generating study materials, and organizing knowledge. \\
        \hline
        \textbf{Personalized learning experience} & Recommend related learning materials.  \\
        \hline
        \textbf{Content creation and generation} & Generate teaching outlines, practice questions, and lesson plans. \\
        \hline
        \textbf{Language learning and teaching} & Provide grammar and vocabulary exercises and enhance their language communication abilities.  \\
        \hline
        \textbf{Cross-language communication and translation}  & Provide real-time translation services.  \\
        \hline
        \textbf{Educational research and data analysis}  & Offer employment prospects, career development paths, and advice on relevant skill development.  \\
        \hline
        \textbf{Virtual experiments and simulations}  & Provide virtual experiment and simulation environments.  \\
        \hline
        \textbf{Career planning and guidance}  & Offer employment prospects, career development paths, and advice.  \\
        \hline
        \textbf{Exam preparation and test-taking support} & Offer practice questions, explanations, and strategies.  \\
        \hline
        \textbf{Academic writing assistance} & Provide guidance on structuring essays, citing sources, refining arguments, and enhancing overall clarity and coherence.  \\
        \hline
        \textbf{Interactive learning experiences} & Create interactive and immersive learning experiences.  \\
        \hline
        \textbf{Lifelong learning and continuing education} & Enable them to acquire new skills, explore new fields, and pursue personal development.  \\
        \hline
    \end{tabularx}
\end{table*}

\textbf{1. Learning assistance tools}: EduLLMs can serve as learning assistance tools, providing support to students in problem-solving, generating study materials, and organizing knowledge. For example, students can ask the model for solution methods to mathematical problems, and the model can generate detailed explanations and step-by-step processes to help students understand and master the concepts.

\textbf{2. Personalized learning experience}: EduLLMs can offer personalized learning content and suggestions based on students' learning needs and interests. For instance, the model can recommend related reading materials, practice questions, and learning resources based on students' learning histories and interests, catering to their individualized requirements.

\textbf{3. Content creation and generation}: EduLLMs can assist educators and content creators in generating educational materials and resources. For example, the models can automatically generate teaching outlines, practice questions, and lesson plans, providing educators with diverse and enriched teaching resources.

\textbf{4. Language learning and teaching}: LLM-empowered education has potential applications in language learning and teaching. For instance, the models can provide grammar and vocabulary exercises to help students improve their language skills. The models can also generate dialogue scenarios for students to practice real-life conversations, enhancing their language communication abilities.

\textbf{5. Cross-language communication and translation}: LLMs can assist in cross-language communication and translation in smart education. For instance, the large models can provide real-time translation services, helping students and educators overcome language barriers and facilitating cross-cultural communication and collaboration.

\textbf{6. Educational research and data analysis}: EduLLMs can analyze extensive educational data (aka educational data mining) \cite{pena2014educational} and provide deep insights and research support. For example, the models can assist researchers in analyzing student's learning behaviors and performances, discovering effective teaching methods and strategies, and providing evidence for educational policy-making.

\textbf{7. Virtual experiments and simulations}: EduLLMs can provide virtual experiment and simulation environments, allowing students to engage in practical experiences. For example, the models can offer virtual chemistry laboratories, enabling students to conduct chemical experiments in safe and controlled environments, honing their practical skills and scientific thinking.

\textbf{8. Career planning and guidance}: EduLLMs provide career planning and guidance to students. For instance, the models can offer employment prospects, career development paths, and advice on relevant skill development based on student's interests, skills, and market demands, assisting students in making informed career planning decisions.

\textbf{9. Exam preparation and test-taking support}: EduLLMs can assist students in preparing for exams and improve their test-taking skills. They can offer practice questions, explanations, and strategies for different types of exams, helping students familiarize themselves with the format, content, and techniques required for successful performance.

\textbf{10. Academic writing assistance}: LLMs can aid students in improving their academic writing skills. They can provide guidance on structuring essays, citing sources, refining arguments, and enhancing overall clarity and coherence. These models can also assist students in developing critical thinking and analytical skills necessary for academic success.

\textbf{11. Interactive learning experiences}: EduLLMs will create interactive and immersive learning experiences. For example, they can simulate historical events, scientific experiments, or virtual field trips, allowing students to engage actively and learn through realistic scenarios. These interactive experiences can enhance student engagement and deepen their understanding of complex concepts.

\textbf{12. Lifelong learning and continuing education}: Educational LLMs can support lifelong learning \cite{li2022massively} and continuing education initiatives. They can provide resources, courses, and learning opportunities for individuals outside traditional educational settings, enabling them to acquire new skills, explore new fields, and pursue personal or professional development at any stage of life.

The versatility of educational LLMs allows for their application across a wide range of educational contexts, from K-12 classrooms to higher education institutions, vocational training, and beyond. By leveraging the capabilities of these models, educational stakeholders can enhance the quality, accessibility, and effectiveness of teaching and learning experiences. In summary, the applications of EduLLMs encompass learning assistance tools, personalized learning experiences, content creation and generation, language learning and teaching, student assignment evaluation, cross-language communication and translation, educational research and data analysis, virtual experiments and simulations, learning content recommendations, and career planning and guidance. These scenarios demonstrate the potential of EduLLMs to provide personalized, efficient, and innovative educational services. However, it is crucial to balance technological advancements with ethical considerations in the application of EduLLMs, ensuring that their usage aligns with educational goals and values while prioritizing individual privacy and data security.

\subsection{Characteristics of Education under LLMs}
\label{sec:characteristics}

Education under large language models (LLMs) exhibits several distinct characteristics, as shown in Fig. \ref{fig:llm}:

\begin{figure}[ht]
    \centering
    \includegraphics[clip,scale=0.35]{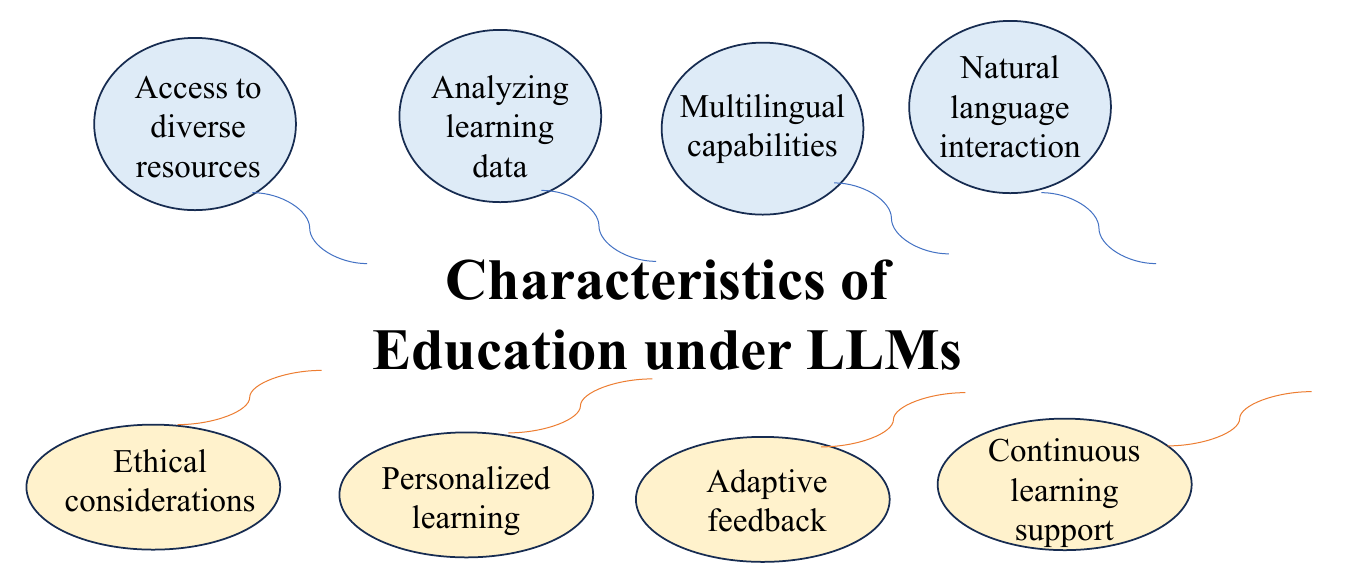}
    \caption{The characteristics of education under LLMs.}
    \label{fig:llm}
\end{figure}

\textbf{1. Personalized learning}: LLMs have the ability to process and analyze vast amounts of data, allowing for personalized learning experiences. They can adapt instructional content, pacing, and assessments to match the unique needs and preferences of individual learners. This personalization enhances the effectiveness and engagement of the learning process.

\textbf{2. Adaptive feedback}: LLMs can provide immediate and adaptive feedback to learners. They can identify areas of weakness or misconceptions and offer tailored explanations and guidance. This real-time feedback helps learners to understand concepts more effectively and make progress at their own pace.

\textbf{3. Access to diverse resources}: For smart education, LLMs have access to a vast amount of information and knowledge. They can provide learners with a wide range of resources, including texts, images, videos, and interactive materials. This access to diverse resources enhances the depth and breadth of learning, enabling learners to explore various perspectives and engage with rich content.

\textbf{4. Natural language interaction}: LLMs are proficient in understanding and generating human language. Learners can engage in natural language conversations with LLMs, asking questions, seeking clarifications, and discussing ideas. This natural language interaction promotes a more conversational and interactive learning experience.

\textbf{5. Continuous learning support}: LLMs can provide continuous learning support beyond traditional classroom hours. Learners can access educational materials, review lessons, and seek assistance from LLMs at any time. Note that this flexibility in learning support accommodates different schedules and learning preferences.

\textbf{6. Content generation and creation}: LLMs can assist in generating educational content. They can automate the creation of quizzes, exercises, and learning materials based on specific learning objectives. This content generation capability reduces the burden on educators and allows for the creation of diverse and customized learning resources.

\textbf{7. Multilingual capabilities}: LLMs are capable of processing and generating content in multiple languages \cite{huang2023not}. This enables learners from different linguistic backgrounds to access educational materials in their native languages, promoting inclusivity and accessibility.

\textbf{8. Analyzing learning data}: Educational LLMs can analyze learning data and provide insights into learners' progress, strengths, and areas for improvement. Educators can utilize these analytics to gain a deeper understanding of learners' learning patterns, adjust instructional strategies, and provide targeted interventions.

\textbf{9. Ethical considerations}: Education under LLMs raises ethical considerations. It is essential to ensure transparency, accountability, and privacy in the use of learner data. Clear guidelines and safeguards should be in place to protect learners' privacy and prevent potential biases or misuse of data.

\textbf{10. Collaboration between humans and LLMs}: LLMs are tools that can enhance and augment human teaching and learning \cite{bernabei2023students}. They are not meant to replace human educators but rather to collaborate with them. Educators can leverage LLMs to provide personalized support, curate content, and facilitate meaningful learning experiences.

\section{Key Points in LLMsEdu}  \label{sec:keys}

\subsection{Training Data and Preprocessing}

Preprocessing steps applied to the data before training may include tokenization, normalization, and other data cleaning techniques. Tokenization involves breaking the text into smaller units, such as words or subwords, to facilitate processing. Normalization may include converting text to lowercase to ensure uniformity and remove case-specific variations. Other cleaning techniques may involve removing irrelevant HTML tags, special characters, or noisy data to enhance the quality of the training data. For educational purposes, when training models to understand and generate text in an educational context, it is crucial to curate datasets that include diverse educational content. This can range from textbooks and scholarly articles to educational websites and forums. These preprocessing steps should be tailored to preserve the educational context, ensuring that the model learns to generate coherent and contextually relevant educational content.

\subsection{Training Process}

Pre-training and fine-tuning play a key role in the construction of educational LLMs. First, in the pre-training stage, the model is initialized through large general text data to achieve the learning of general language features such as syntax, semantics, and logical relationships. This provides the model with broad language understanding capabilities, allowing it to understand and process a variety of language tasks. Next, in the fine-tuning phase, fine-tuning is performed by collecting domain-specific data according to specific task requirements in the education field. This ensures that the model can better adapt to the tasks and show superior performance in the education field. During the fine-tuning process, pre-trained model weights are used for initialization, which provides a strong foundation for the model to learn specific tasks. Adjust model parameters through supervised learning to adapt them to the specific requirements of the task, and ensure that the model reaches a satisfactory level on educational tasks through performance evaluation. Hyperparameter tuning further optimizes model performance, such as by adjusting the learning rate and batch size. Ultimately, by saving the fine-tuned model, it becomes a powerful tool that can be deployed and applied to specific educational tasks. Therefore, the entire training process enables the model to achieve excellent results in a wide range of language understanding and specific educational tasks, providing a powerful language processing tool for smart education.

\subsection{Integration with Educational Technologies}

Finally, they can be seamlessly integrated into various practical applications within educational technology to enhance the overall learning experience. LLMs can power chatbots, providing personalized support by addressing queries related to course content, assignments, or general information, with the added advantage of 24/7 availability. LLMs can be incorporated into intelligent tutoring systems, delivering personalized learning experiences by offering customized guidance and recommendations to students. They can also automate the generation of educational content, including quizzes, tests, and study materials, thereby saving educators valuable time. Moreover, LLMs have applications on language learning platforms, facilitating conversational practice through realistic dialogue simulations and offering real-time feedback on grammar usage. These technologies can extend to virtual labs and simulations, enhancing students' practical learning experiences through natural language interactions. Overall, the application of LLMs in educational technology necessitates considerations for ethical issues, data privacy, and potential biases in the models. Continuous user feedback and improvement are crucial for optimizing learning outcomes.

\section{Challenges and Future Directions} 
\subsection{Challenges and Issues}
\label{sec:challenges}

The application of LLMs for education brings forth numerous potential challenges and issues. Here are 10 possible challenges related to LLM4Edu, along with detailed descriptions:

\textbf{1. Privacy protection} \cite{chen2023privacy,gan2018privacy}: In general, EduLLMs deal with a vast amount of student data, including personal information, learning records, and behavioral data. This raises concerns regarding privacy protection. Ensuring the security and privacy of student data becomes a significant challenge, necessitating rigorous data security measures and privacy policies to safeguard student rights.

\textbf{2. Data bias}: The data used during the training process of EduLLMs may contain biases, which can result in biased outputs from the models \cite{schramowski2022large}. For instance, if there are biases in the training data concerning gender or race, the models may reflect these biases and have unfair effects on students. Eliminating data bias is an important challenge to ensure the fairness and reliability of the models.

\textbf{3. Algorithm transparency}: EduLLMs often consist of complex neural network models, and their decision-making processes can be difficult to interpret and understand. Algorithm transparency refers to the extent to which the model's decision-making process can be explained and understood \cite{rader2018explanations}. In education, students and teachers need to understand how the models make recommendations and evaluations to trust and utilize them.

\textbf{4. Technical feasibility}: Educational LLMs typically require substantial computational resources and storage space for training and inference. In certain educational environments, especially in resource-constrained schools or regions, these requirements may not be met. Hence, ensuring the technical feasibility of EduLLMs to operate reliably in various educational settings is a critical challenge.

\textbf{5. Human interaction and emotion}: Education involves rich human interactions and emotional experiences. EduLLMs still face challenges in simulating human teacher-student interactions. For example, in terms of emotion analysis, models may struggle to accurately understand students' emotional states and provide appropriate support \cite{aldrup2022empathy}. Addressing these challenges, especially in the Metaverse \cite{sun2022metaverse,yang2023human}, requires further research and technological innovation.

\textbf{6. Accessibility}: The application of EduLLMs should have broad accessibility to meet the needs of diverse learners. This includes support for students with disabilities, such as assistive features for visually and hearing-impaired students. Ensuring that accessibility needs are considered in the design and implementation of EduLLMs is a significant challenge.

\textbf{7. Credibility and quality assessment}: Ensuring the credibility and quality assessment of EduLLMs is crucial. Students and teachers need to have confidence that the recommendations and feedback provided by the models are accurate and reliable \cite{boud2013rethinking}. Therefore, conducting credibility and quality assessments of EduLLMs is an important challenge. This involves establishing evaluation criteria and metrics to validate the model's performance and effectiveness while ensuring its reliability in educational practice.

\textbf{8. Teacher roles and professional development}: The use of EduLLMs may impact teacher roles and professional development. Firstly, EduLLMs can provide instructional assistance and personalized learning support, alleviating teachers' workload. Secondly, teachers need to adapt to and master the technologies and tools related to EduLLMs to collaborate and work effectively with them. This presents new requirements and challenges for teacher professional development.

\subsection{Future Directions}
\label{sec:directions}

Here are some possible research directions for EduLLMs in the future, along with detailed descriptions:

\textbf{1. Model interpretability}: Educational LLMs often consist of complex neural network structures, and their decision-making processes can be difficult to interpret and understand. To establish the credibility and acceptability of EduLLMs, further research is challenging on how to explain the model's decision-making process, enabling teachers, students, and other stakeholders to comprehend and trust the model's recommendations and evaluations.

\textbf{2. Personalized learning support}: One major application of EduLLMs is to provide personalized learning support. Future research can explore how to better utilize models to understand students' learning needs, interests, and learning styles, in order to offer more accurate and personalized learning suggestions and resources.

\textbf{3. Emotional intelligence}: Education involves emotional factors such as students' emotional states and experiences. Future research can focus on integrating emotional intelligence into EduLLMs, enabling the models to accurately recognize and understand students' emotional states and provide appropriate emotional support and guidance when needed.

\textbf{4. Evaluation and assessment}: Evaluating the effectiveness and impact of EduLLMs is important. Future research can focus on establishing effective evaluation methods and metrics to assess the influence of EduLLMs on students' learning outcomes, learning processes, and learning experiences.

\textbf{5. Social equity}: The application of EduLLMs in providing personalized learning may raise issues of social equity. Future research can explore how to address these issues through the design and implementation of models, ensuring that their applications do not exacerbate educational inequalities but instead promote a fair and inclusive learning environment.

\textbf{6. Educational ethics}: The application of EduLLMs raises ethical issues such as privacy protection, data usage, and the model's moral responsibility. Future research can focus on establishing appropriate ethical guidelines and frameworks to guide the development, use, and evaluation of EduLLMs.

\textbf{7. Cross-cultural adaptability}: The research and application of EduLLMs need to consider the needs and differences of learners from different cultures and backgrounds. Future research can focus on making EduLLMs cross-culturally adaptable to better meet the needs of learners worldwide.

\textbf{8. Long-term learning and development}: Research on EduLLMs should not only focus on short-term effects during the learning process but also consider students' long-term learning and development. Future research can explore how EduLLMs can support students' long-term learning goals, facilitate continuous growth, and promote lifelong learning.

\section{Conclusion} \label{sec:conclusion}

The application of LLMs in the field of education has broad prospects. This review provides a systematic summary and analysis of the research background, motivation, and application of educational large models. It first introduces the research background and motivation of LLMs and explains the essence of large models. It then discusses the relationship between intelligent education and educational LLMs, and summarizes the current research status of educational LLMs. Finally, by reviewing existing research, this article provides guidance and insights for educators, researchers, and policy-makers to gain a deep understanding of the potential opportunities and challenges of educational LLMs, and provides guidance for further advancing the development and application of educational LLMs. However, the development and applications of educational LLMs still face technical, ethical, and practical challenges, requiring further research and exploration. 

With the advancement of technology and the evolution of educational needs, educational large models will play an increasingly important role in providing more efficient and personalized support and services for education. We believe that AI-driven education is one of the most innovative and forward-looking directions in the field of education today. It can be foreseen that in the future, with the continuous development and improvement of artificial intelligence, the future of smart education will be more digitalized and humanized, as well as more diverse and personalized.

\section*{Acknowledgment}

This research was supported in part by the National Natural Science Foundation of China (Nos. 62002136 and 62272196), Natural Science Foundation of Guangdong Province (No. 2022A1515011861), Fundamental Research Funds for the Central Universities of Jinan University (No. 21622416), the Young Scholar Program of Pazhou Lab (No. PZL2021KF0023), Engineering Research Center of Trustworthy AI, Ministry of Education (Jinan University), and Guangdong Key Laboratory of Data Security and Privacy Preserving. Dr. Wensheng Gan is the corresponding author of this paper.

\bibliographystyle{IEEEtran}
\bibliography{llmedu.bib}

\begin{thebibliography}{10}
\providecommand{\url}[1]{#1}
\csname url@samestyle\endcsname
\providecommand{\newblock}{\relax}
\providecommand{\bibinfo}[2]{#2}
\providecommand{\BIBentrySTDinterwordspacing}{\spaceskip=0pt\relax}
\providecommand{\BIBentryALTinterwordstretchfactor}{4}
\providecommand{\BIBentryALTinterwordspacing}{\spaceskip=\fontdimen2\font plus
\BIBentryALTinterwordstretchfactor\fontdimen3\font minus
  \fontdimen4\font\relax}
\providecommand{\BIBforeignlanguage}[2]{{%
\expandafter\ifx\csname l@#1\endcsname\relax
\typeout{** WARNING: IEEEtran.bst: No hyphenation pattern has been}%
\typeout{** loaded for the language `#1'. Using the pattern for}%
\typeout{** the default language instead.}%
\else
\language=\csname l@#1\endcsname
\fi
#2}}
\providecommand{\BIBdecl}{\relax}
\BIBdecl

\bibitem{sun2022big}
J.~Sun, W.~Gan, Z.~Chen, J.~Li, and P.~S. Yu, ``Big data meets {Metaverse}: A
  survey,'' \emph{arXiv preprint arXiv:2210.16282}, 2022.

\bibitem{sun2023internet}
J.~Sun, W.~Gan, H.~Chao, P.~S. Yu, and W.~Ding, ``Internet of behaviors: A
  survey,'' \emph{IEEE Internet of Things Journal}, vol.~10, no.~13, pp.
  11\,117--11\,134, 2023.

\bibitem{wan2023web3}
S.~Wan, H.~Lin, W.~Gan, J.~Chen, and P.~S. Yu, ``Web3: The next internet
  revolution,'' \emph{arXiv preprint, arXiv:2304.06111}, 2023.

\bibitem{gan2023web}
W.~Gan, Z.~Ye, S.~Wan, and P.~S. Yu, ``Web 3.0: The future of internet,'' in
  \emph{Companion Proceedings of the Web Conference}.\hskip 1em plus 0.5em
  minus 0.4em\relax ACM, 2023, pp. 1266--1275.

\bibitem{zhao2023survey}
W.~X. Zhao, K.~Zhou, J.~Li, T.~Tang, X.~Wang, Y.~Hou, Y.~Min, B.~Zhang,
  J.~Zhang, Z.~Dong \emph{et~al.}, ``A survey of large language models,''
  \emph{arXiv preprint, arXiv:2303.18223}, 2023.

\bibitem{gan2023large}
W.~Gan, Z.~Qi, J.~Wu, and J.~C.~W. Lin, ``Large language models in education:
  Vision and opportunities,'' in \emph{IEEE International Conference on Big
  Data}.\hskip 1em plus 0.5em minus 0.4em\relax IEEE, 2023, pp. 1--10.

\bibitem{gan2023model}
W.~Gan, S.~Wan, and P.~S. Yu, ``Model-as-a-service ({MaaS}): A survey,'' in
  \emph{IEEE International Conference on Big Data}.\hskip 1em plus 0.5em minus
  0.4em\relax IEEE, 2023, pp. 1--10.

\bibitem{zeng2023large}
F.~Zeng, W.~Gan, Y.~Wang, N.~Liu, and P.~S. Yu, ``Large language models for
  robotics: A survey,'' \emph{arXiv preprint, arXiv:2311.07226}, 2023.

\bibitem{wu2023ai}
J.~Wu, W.~Gan, Z.~Chen, S.~Wan, and H.~Lin, ``{AI}-generated content ({AIGC}):
  A survey,'' \emph{arXiv preprint, arXiv:2304.06632}, 2023.

\bibitem{xiao2023survey}
Y.~Xiao, L.~Wu, J.~Guo, J.~Li, M.~Zhang, T.~Qin, and T.~Liu, ``A survey on
  non-autoregressive generation for neural machine translation and beyond,''
  \emph{IEEE Transactions on Pattern Analysis and Machine Intelligence},
  vol.~45, pp. 11\,407--11\,427, 2023.

\bibitem{ziems2021moral}
C.~Ziems, J.~Yu, Y.-C. Wang, A.~Halevy, and D.~Yang, ``The moral integrity
  corpus: A benchmark for ethical dialogue systems,'' in \emph{The 60th Annual
  Meeting of the Association for Computational Linguistics}, 2022, pp.
  3755--3773.

\bibitem{aldhafeeri2022effectiveness}
F.~M. Aldhafeeri and A.~A. Alotaibi, ``Effectiveness of digital education
  shifting model on high school students’ engagement,'' \emph{Education and
  Information Technologies}, vol.~27, no.~5, pp. 6869--6891, 2022.

\bibitem{lin2022metaverse}
H.~Lin, S.~Wan, W.~Gan, J.~Chen, and H.~Chao, ``Metaverse in education: Vision,
  opportunities, and challenges,'' in \emph{IEEE International Conference on
  Big Data}.\hskip 1em plus 0.5em minus 0.4em\relax IEEE, 2022, pp. 2857--2866.

\bibitem{mcnamara2023istart}
D.~S. McNamara, T.~Arner, R.~Butterfuss, Y.~Fang, M.~Watanabe, N.~Newton, K.~S.
  McCarthy, L.~K. Allen, and R.~D. Roscoe, ``{iSTART}: Adaptive comprehension
  strategy training and stealth literacy assessment,'' \emph{International
  Journal of Human--Computer Interaction}, vol.~39, no.~11, pp. 2239--2252,
  2023.

\bibitem{kristianto2023offline}
H.~Kristianto and L.~Gandajaya, ``Offline vs online problem-based learning: A
  case study of student engagement and learning outcomes,'' \emph{Interactive
  Technology and Smart Education}, vol.~20, no.~1, pp. 106--121, 2023.

\bibitem{smith2016widening}
P.~S. Smith, P.~J. Trygstad, and E.~R. Banilower, ``Widening the gap: Unequal
  distribution of resources for {K}-12 science instruction.'' \emph{Education
  Policy Analysis Archives}, vol.~24, no.~8, p.~n8, 2016.

\bibitem{piety2014educational}
P.~J. Piety, D.~T. Hickey, and M.~Bishop, ``Educational data sciences: Framing
  emergent practices for analytics of learning, organizations, and systems,''
  in \emph{The Fourth International Conference on Learning Analytics and
  Knowledge}, 2014, pp. 193--202.

\bibitem{vermunt2017learning}
J.~D. Vermunt and V.~Donche, ``A learning patterns perspective on student
  learning in higher education: state of the art and moving forward,''
  \emph{Educational Psychology Review}, vol.~29, pp. 269--299, 2017.

\bibitem{aziz2012evaluation}
A.~A. Aziz, K.~M. Yusof, and J.~M. Yatim, ``Evaluation on the effectiveness of
  learning outcomes from students’ perspectives,'' \emph{Procedia-Social and
  Behavioral Sciences}, vol.~56, pp. 22--30, 2012.

\bibitem{fang2021personalized}
C.~Fang and Q.~Lu, ``Personalized recommendation model of high-quality
  education resources for college students based on data mining,''
  \emph{Complexity}, vol. 2021, pp. 1--11, 2021.

\bibitem{bhargava2022commonsense}
P.~Bhargava and V.~Ng, ``Commonsense knowledge reasoning and generation with
  pre-trained language models: A survey,'' in \emph{The AAAI Conference on
  Artificial Intelligence}, 2022, pp. 12\,317--12\,325.

\bibitem{kasneci2023chatgpt}
E.~Kasneci, K.~Se{\ss}ler, S.~K{\"u}chemann, M.~Bannert, D.~Dementieva,
  F.~Fischer, U.~Gasser, G.~Groh, S.~G{\"u}nnemann, E.~H{\"u}llermeier
  \emph{et~al.}, ``{ChatGPT} for good? on opportunities and challenges of large
  language models for education,'' \emph{Learning and Individual Differences},
  vol. 103, p. 102274, 2023.

\bibitem{raj2022systematic}
N.~S. Raj and V.~Renumol, ``A systematic literature review on adaptive content
  recommenders in personalized learning environments from 2015 to 2020,''
  \emph{Journal of Computers in Education}, vol.~9, no.~1, pp. 113--148, 2022.

\bibitem{wang2023unified}
Z.~Wang, W.~Yan, C.~Zeng, Y.~Tian, S.~Dong \emph{et~al.}, ``A unified
  interpretable intelligent learning diagnosis framework for learning
  performance prediction in intelligent tutoring systems,'' \emph{International
  Journal of Intelligent Systems}, vol. 2023, 2023.

\bibitem{rudolph2023chatgpt}
J.~Rudolph, S.~Tan, and S.~Tan, ``{ChatGPT}: Bullshit spewer or the end of
  traditional assessments in higher education?'' \emph{Journal of Applied
  Learning and Teaching}, vol.~6, no.~1, 2023.

\bibitem{marshall2022implementing}
R.~Marshall, A.~Pardo, D.~Smith, and T.~Watson, ``Implementing next generation
  privacy and ethics research in education technology,'' \emph{British Journal
  of Educational Technology}, vol.~53, no.~4, pp. 737--755, 2022.

\bibitem{kizilcec2022algorithmic}
R.~F. Kizilcec and H.~Lee, ``Algorithmic fairness in education,'' in \emph{The
  Ethics of Artificial Intelligence in Education}, 2022, pp. 174--202.

\bibitem{lee2023rise}
H.~Lee, ``The rise of {ChatGPT}: Exploring its potential in medical
  education,'' \emph{Anatomical Sciences Education}, 2023.

\bibitem{zachary2022mentor}
L.~J. Zachary and L.~Z. Fain, \emph{The mentor's guide: Facilitating effective
  learning relationships}.\hskip 1em plus 0.5em minus 0.4em\relax John Wiley \&
  Sons, 2022.

\bibitem{shunkov2022prospective}
V.~Shunkov, O.~Shevtsova, V.~Koval, T.~Grygorenko, L.~Yefymenko, Y.~Smolianko,
  and O.~Kuchai, ``Prospective directions of using multimedia technologies in
  the training of future specialists,'' 2022.

\bibitem{tang2023science}
R.~Tang, Y.-N. Chuang, and X.~Hu, ``The science of detecting {LLM}-generated
  texts,'' \emph{arXiv preprint, arXiv:2303.07205}, 2023.

\bibitem{baidoo2023education}
D.~Baidoo-Anu and L.~O. Ansah, ``Education in the era of generative artificial
  intelligence ({AI}): Understanding the potential benefits of {ChatGPT} in
  promoting teaching and learning,'' \emph{Journal of AI}, vol.~7, no.~1, pp.
  52--62, 2023.

\bibitem{weissweiler2022better}
L.~Weissweiler, V.~Hofmann, A.~K{\"o}ksal, and H.~Sch{\"u}tze, ``The better
  your syntax, the better your semantics? probing pretrained language models
  for the english comparative correlative,'' \emph{arXiv preprint,
  arXiv:2210.13181}, 2022.

\bibitem{meng2022generating}
Y.~Meng, J.~Huang, Y.~Zhang, and J.~Han, ``Generating training data with
  language models: Towards zero-shot language understanding,'' \emph{Advances
  in Neural Information Processing Systems}, vol.~35, pp. 462--477, 2022.

\bibitem{agarwal2022chatbots}
S.~Agarwal, B.~Agarwal, and R.~Gupta, ``Chatbots and virtual assistants: a
  bibliometric analysis,'' \emph{Library Hi Tech}, vol.~40, no.~4, pp.
  1013--1030, 2022.

\bibitem{gao2022impact}
J.~Gao, L.~Ren, Y.~Yang, D.~Zhang, and L.~Li, ``The impact of artificial
  intelligence technology stimuli on smart customer experience and the
  moderating effect of technology readiness,'' \emph{International Journal of
  Emerging Markets}, vol.~17, no.~4, pp. 1123--1142, 2022.

\bibitem{salvagno2023can}
M.~Salvagno, F.~S. Taccone, A.~G. Gerli \emph{et~al.}, ``Can artificial
  intelligence help for scientific writing?'' \emph{Critical Care}, vol.~27,
  no.~1, pp. 1--5, 2023.

\bibitem{ertuugrul2023lifelong}
U.~Ertu{\u{g}}rul, ``Lifelong learning motivation scale ({LLMs}): Validity and
  reliability study,'' \emph{Journal of Teacher Education and Lifelong
  Learning}, vol.~5, no.~1, pp. 429--438, 2023.

\bibitem{vaswani2017attention}
A.~Vaswani, N.~Shazeer, N.~Parmar, J.~Uszkoreit, L.~Jones, A.~N. Gomez,
  {\L}.~Kaiser, and I.~Polosukhin, ``Attention is all you need,''
  \emph{Advances in Neural Information Processing Systems}, vol.~30, 2017.

\bibitem{shaw2018self}
P.~Shaw, J.~Uszkoreit, and A.~Vaswani, ``Self-attention with relative position
  representations,'' in \emph{NAACL-HLT}, 2018, pp. 464--468.

\bibitem{zoph2020rethinking}
B.~Zoph, G.~Ghiasi, T.-Y. Lin, Y.~Cui, H.~Liu, E.~D. Cubuk, and Q.~Le,
  ``Rethinking pre-training and self-training,'' \emph{Advances in Neural
  Information Processing Systems}, vol.~33, pp. 3833--3845, 2020.

\bibitem{howard2018universal}
J.~Howard and S.~Ruder, ``Universal language model fine-tuning for text
  classification,'' in \emph{the 56th Annual Meeting of the Association for
  Computational Linguistics}, 2018, pp. 328--339.

\bibitem{kandpal2023large}
N.~Kandpal, H.~Deng, A.~Roberts, E.~Wallace, and C.~Raffel, ``Large language
  models struggle to learn long-tail knowledge,'' in \emph{International
  Conference on Machine Learning}.\hskip 1em plus 0.5em minus 0.4em\relax PMLR,
  2023, pp. 15\,696--15\,707.

\bibitem{zeng2023distributed}
F.~Zeng, W.~Gan, Y.~Wang, and P.~S. Yu, ``Distributed training of large
  language models,'' in \emph{The 29th IEEE International Conference on
  Parallel and Distributed Systems}.\hskip 1em plus 0.5em minus 0.4em\relax
  IEEE, 2023, pp. 1--8.

\bibitem{jin2023convergence}
B.~Jin and {\v{Z}}.~Kereta, ``On the convergence of stochastic gradient descent
  for linear inverse problems in banach spaces,'' \emph{SIAM Journal on Imaging
  Sciences}, vol.~16, no.~2, pp. 671--705, 2023.

\bibitem{reyad2023modified}
M.~Reyad, A.~M. Sarhan, and M.~Arafa, ``A modified adam algorithm for deep
  neural network optimization,'' \emph{Neural Computing and Applications}, pp.
  1--18, 2023.

\bibitem{kang2023scaling}
M.~Kang, J.-Y. Zhu, R.~Zhang, J.~Park, E.~Shechtman, S.~Paris, and T.~Park,
  ``Scaling up gans for text-to-image synthesis,'' in \emph{The IEEE/CVF
  Conference on Computer Vision and Pattern Recognition}, 2023, pp.
  10\,124--10\,134.

\bibitem{zhu2023unsupervised}
P.~Zhu, X.~Hou, K.~Tang, Y.~Liu, Y.-P. Zhao, and Z.~Wang, ``Unsupervised
  feature selection through combining graph learning and L2, 0-norm
  constraint,'' \emph{Information Sciences}, vol. 622, pp. 68--82, 2023.

\bibitem{xu2023survey}
C.~Xu and J.~McAuley, ``A survey on model compression and acceleration for
  pretrained language models,'' in \emph{The AAAI Conference on Artificial
  Intelligence}, vol.~37, no.~9, 2023, pp. 10\,566--10\,575.

\bibitem{bajaj2018smart}
R.~Bajaj and V.~Sharma, ``Smart education with artificial intelligence based
  determination of learning styles,'' \emph{Procedia Computer Science}, vol.
  132, pp. 834--842, 2018.

\bibitem{hampel2001steam}
T.~Hampel and R.~Keil-Slawik, ``steam: structuring information in
  team-distributed knowledge management in cooperative learning environments,''
  \emph{Journal on Educational Resources in Computing}, vol.~1, no. 2es, pp.
  3--es, 2001.

\bibitem{chen2022metaverse}
Z.~Chen, J.~Wu, W.~Gan, and Z.~Qi, ``Metaverse security and privacy: An
  overview,'' in \emph{IEEE International Conference on Big Data}.\hskip 1em
  plus 0.5em minus 0.4em\relax IEEE, 2022, pp. 2950--2959.

\bibitem{chen2023privacy}
Y.~Chen, W.~Gan, Y.~Wu, and P.~S. Yu, ``Privacy-preserving federated mining of
  frequent itemsets,'' \emph{Information Sciences}, vol. 625, pp. 504--520,
  2023.

\bibitem{may2011using}
M.~May and S.~George, ``Using students' tracking data in e-learning: Are we
  always aware of security and privacy concerns?'' in \emph{The IEEE 3rd
  International Conference on Communication Software and Networks}.\hskip 1em
  plus 0.5em minus 0.4em\relax IEEE, 2011, pp. 10--14.

\bibitem{fournier2022pattern}
P.~Fournier-Viger, W.~Gan, Y.~Wu, M.~Nouioua, W.~Song, T.~Truong, and H.~Duong,
  ``Pattern mining: Current challenges and opportunities,'' in
  \emph{International Conference on Database Systems for Advanced
  Applications}.\hskip 1em plus 0.5em minus 0.4em\relax Springer, 2022, pp.
  34--49.

\bibitem{gan2019survey}
W.~Gan, J.~C.~W. Lin, P.~Fournier-Viger, H.~C. Chao, and P.~S. Yu, ``A survey
  of parallel sequential pattern mining,'' \emph{ACM Transactions on Knowledge
  Discovery from Data}, vol.~13, no.~3, pp. 1--34, 2019.

\bibitem{herodotou2019large}
C.~Herodotou, B.~Rienties, A.~Boroowa, Z.~Zdrahal, and M.~Hlosta, ``A
  large-scale implementation of predictive learning analytics in higher
  education: The teachers’ role and perspective,'' \emph{Educational
  Technology Research and Development}, vol.~67, pp. 1273--1306, 2019.

\bibitem{filgueiras2023artificial}
F.~Filgueiras, ``Artificial intelligence and education governance,''
  \emph{Education, Citizenship and Social Justice}, p. 17461979231160674, 2023.

\bibitem{al2023survey}
T.~A. Al-Qablan, M.~H. Mohd~Noor, M.~A. Al-Betar, and A.~T. Khader, ``A survey
  on sentiment analysis and its applications,'' \emph{Neural Computing and
  Applications}, pp. 1--35, 2023.

\bibitem{jordan2015machine}
M.~I. Jordan and T.~M. Mitchell, ``Machine learning: Trends, perspectives, and
  prospects,'' \emph{Science}, vol. 349, no. 6245, pp. 255--260, 2015.

\bibitem{lecun2015deep}
Y.~LeCun, Y.~Bengio, and G.~Hinton, ``Deep learning,'' \emph{Nature}, vol. 521,
  no. 7553, pp. 436--444, 2015.

\bibitem{kaelbling1996reinforcement}
L.~P. Kaelbling, M.~L. Littman, and A.~W. Moore, ``Reinforcement learning: A
  survey,'' \emph{Journal of Artificial Intelligence Research}, vol.~4, pp.
  237--285, 1996.

\bibitem{carta2023grounding}
T.~Carta, C.~Romac, T.~Wolf, S.~Lamprier, O.~Sigaud, and P.-Y. Oudeyer,
  ``Grounding large language models in interactive environments with online
  reinforcement learning,'' \emph{arXiv preprint, arXiv:2302.02662}, 2023.

\bibitem{gan2020huopm}
W.~Gan, J.~C.-W. Lin, P.~Fournier-Viger, H.~Chao, and P.~S. Yu, ``{HUOPM}:
  High-utility occupancy pattern mining,'' \emph{IEEE Transactions on
  Cybernetics}, vol.~50, no.~3, pp. 1195--1208, 2020.

\bibitem{gan2021survey}
W.~Gan, J.~C.-W. Lin, P.~Fournier-Viger, H.~Chao, V.~S. Tseng, and P.~S. Yu,
  ``A survey of utility-oriented pattern mining,'' \emph{IEEE Transactions on
  Knowledge and Data Engineering}, vol.~33, no.~4, pp. 1306--1327, 2021.

\bibitem{thomas2017predicting}
C.~Thomas and D.~B. Jayagopi, ``Predicting student engagement in classrooms
  using facial behavioral cues,'' in \emph{The 1st ACM SIGCHI Workshop on
  Multimodal Interaction for Education}, 2017, pp. 33--40.

\bibitem{wong2022leveraging}
A.~B. Wong, Z.~Huang, and K.~Wu, ``Leveraging audible and inaudible signals for
  pronunciation training by sensing articulation through a smartphone,''
  \emph{Speech Communication}, vol. 144, pp. 42--56, 2022.

\bibitem{wu2023multimodal}
J.~Wu, W.~Gan, Z.~Chen, S.~Wan, and P.~S. Yu, ``Multimodal large language
  models: A survey,'' in \emph{IEEE International Conference on Big
  Data}.\hskip 1em plus 0.5em minus 0.4em\relax IEEE, 2023, pp. 1--10.

\bibitem{martinez2020data}
R.~Martinez-Maldonado, V.~Echeverria, G.~Fernandez~Nieto, and
  S.~Buckingham~Shum, ``From data to insights: A layered storytelling approach
  for multimodal learning analytics,'' in \emph{The CHI Conference on Human
  Factors in Computing Systems}, 2020, pp. 1--15.

\bibitem{li2023prompt}
L.~Li, Y.~Zhang, and L.~Chen, ``Prompt distillation for efficient {LLM}-based
  recommendation,'' in \emph{The 32nd ACM International Conference on
  Information and Knowledge Management}, 2023, pp. 1348--1357.

\bibitem{pena2014educational}
A.~Pe{\~n}a-Ayala, ``Educational data mining: A survey and a data mining-based
  analysis of recent works,'' \emph{Expert Systems with Applications}, vol.~41,
  no.~4, pp. 1432--1462, 2014.

\bibitem{li2022massively}
B.~Li, R.~Pang, Y.~Zhang, T.~N. Sainath, T.~Strohman, P.~Haghani, Y.~Zhu,
  B.~Farris, N.~Gaur, and M.~Prasad, ``Massively multilingual asr: A lifelong
  learning solution,'' in \emph{IEEE International Conference on Acoustics,
  Speech and Signal Processing}.\hskip 1em plus 0.5em minus 0.4em\relax IEEE,
  2022, pp. 6397--6401.

\bibitem{huang2023not}
H.~Huang, T.~Tang, D.~Zhang, W.~X. Zhao, T.~Song, Y.~Xia, and F.~Wei, ``Not all
  languages are created equal in {LLMs}: Improving multilingual capability by
  cross-lingual-thought prompting,'' \emph{arXiv preprint, arXiv:2305.07004},
  2023.

\bibitem{bernabei2023students}
M.~Bernabei, S.~Colabianchi, A.~Falegnami, and F.~Costantino, ``Students’ use
  of large language models in engineering education: A case study on technology
  acceptance, perceptions, efficacy, and detection chances,'' \emph{Computers
  and Education: Artificial Intelligence}, p. 100172, 2023.

\bibitem{gan2018privacy}
W.~Gan, C.-W.~J. Lin, H.~C. Chao, S.~L. Wang, and P.~S. Yu, ``Privacy
  preserving utility mining: a survey,'' in \emph{IEEE International Conference
  on Big Data}.\hskip 1em plus 0.5em minus 0.4em\relax IEEE, 2018, pp.
  2617--2626.

\bibitem{schramowski2022large}
P.~Schramowski, C.~Turan, N.~Andersen, C.~A. Rothkopf, and K.~Kersting, ``Large
  pre-trained language models contain human-like biases of what is right and
  wrong to do,'' \emph{Nature Machine Intelligence}, vol.~4, no.~3, pp.
  258--268, 2022.

\bibitem{rader2018explanations}
E.~Rader, K.~Cotter, and J.~Cho, ``Explanations as mechanisms for supporting
  algorithmic transparency,'' in \emph{The CHI Conference on Human Factors in
  Computing Systems}, 2018, pp. 1--13.

\bibitem{aldrup2022empathy}
K.~Aldrup, B.~Carstensen, and U.~Klusmann, ``Is empathy the key to effective
  teaching? a systematic review of its association with teacher-student
  interactions and student outcomes,'' \emph{Educational {Psychology}
  {Review}}, vol.~34, no.~3, pp. 1177--1216, 2022.

\bibitem{sun2022metaverse}
J.~Sun, W.~Gan, H.~Chao, and P.~S. Yu, ``Metaverse: Survey, applications,
  security, and opportunities,'' \emph{arXiv preprint arXiv:2210.07990}, 2022.

\bibitem{yang2023human}
R.~Yang, L.~Li, W.~Gan, Z.~Chen, and Z.~Qi, ``The human-centric {Metaverse}: A
  survey,'' in \emph{Companion Proceedings of the ACM Web Conference}, 2023,
  pp. 1296--1306.

\bibitem{boud2013rethinking}
D.~Boud and E.~Molloy, ``Rethinking models of feedback for learning: the
  challenge of design,'' \emph{Assessment \& Evaluation in Higher Education},
  vol.~38, no.~6, pp. 698--712, 2013.

\end{thebibliography}

\end{document}